\documentclass[journal]{IEEEtran}

\PassOptionsToPackage{dvipsnames}{xcolor}

\usepackage{cite}
\usepackage{algorithmic}
\usepackage{textcomp}

\usepackage{multirow}
\usepackage{lipsum}
\usepackage{amsmath,amssymb,amsfonts}
\usepackage[ruled,vlined]{algorithm2e}
\usepackage{graphicx,color}
\graphicspath{{./Figures/}}
\usepackage{capt-of} 

\usepackage[hidelinks]{hyperref}
\usepackage{pgfplots}
\usepgfplotslibrary{groupplots}
\pgfplotsset{width=5cm,compat=1.9} 

\usepackage{csquotes}
\usepackage{siunitx}
\usepackage[dvipsnames]{xcolor}
\usepackage{balance}
\usepackage{svg}
\usepackage{orcidlink}
\usepackage{bookmark}

\usepackage[colorinlistoftodos,textsize=tiny,disable]{todonotes}

\def\BibTeX{{\rm B\kern-.05em{\sc i\kern-.025em b}\kern-.08em
    T\kern-.1667em\lower.7ex\hbox{E}\kern-.125emX}}

\begin{document}	

\title{Understanding Autonomous Driving Datasets by Describing Differences between Image Subsets in Natural Language}

\author{Julian~Truetsch\IEEEauthorrefmark{1}\IEEEauthorrefmark{2},
        Felix~Hauser\IEEEauthorrefmark{1}\IEEEauthorrefmark{2},
        Christoph~Stiller\IEEEauthorrefmark{2},~\IEEEmembership{Fellow,~IEEE},
        and~Frank~Bieder\IEEEauthorrefmark{1}\IEEEauthorrefmark{2}%
\thanks{\IEEEauthorrefmark{1}FZI Research Center for Information Technology, 76131 Karlsruhe, Germany.}%
\thanks{\IEEEauthorrefmark{2}Karlsruhe Institute of Technology (KIT), 76131 Karlsruhe, Germany.}%
\thanks{Corresponding author: Julian Truetsch (e-mail: truetsch@fzi.de).}%
\thanks{The research leading to these results is funded by the German Federal Ministry for Economic Affairs and Energy within the projects jbDATA and nxtAIM.
    We acknowledge support by the Karlsruhe School of Optics~\& Photonics (KSOP) and the Ministry of Science, Research and Arts of Baden-Württemberg as part of the sustainability financing of the projects of the Excellence Initiative~II.
    The authors gratefully acknowledge the computing time provided on the high-performance computer HoreKa by the National High-Performance Computing Center at KIT (NHR@KIT). This center is jointly supported by the Federal Ministry of Education and Research and the Ministry of Science, Research and Arts of Baden-Württemberg, as part of the \href{https://www.nhr-verein.de/en/our-partners}{National High-Performance Computing (NHR) joint funding program}. HoreKa is partly funded by the German Research Foundation (DFG).}%
\thanks{This work has been submitted to the IEEE for possible publication. Copyright may be transferred without notice, after which this version may no longer be accessible.}}

\markboth{Preprint}%
{Truetsch \MakeLowercase{\textit{et al.}}: Understanding Autonomous Driving Datasets by Describing Differences between Image Subsets}

\maketitle

\begin{abstract}
Understanding the composition of large-scale autonomous driving datasets is essential for safety, robustness, and reliable operation across domains.
For example, domain shift between locations could lead to the operating environment being misaligned with the training data, resulting in potentially dangerous performance degradation.
Yet, existing data analysis pipelines largely rely on metadata, predefined labels, or manual inspection, which provide limited semantic insight or do not scale.
This paper studies \emph{set difference captioning}: given two subsets of images, the goal is to produce a natural-language hypothesis describing differences between the target and reference set.
Building on a two-stage formulation, we adapt the method to autonomous driving by focusing on object-centric patches derived from object detection, which simplifies aggregation and enables attribution of differences to specific object instances or categories.
To evaluate this setting in-domain, we introduce a new benchmark, AD-Diff Bench.
Low-\emph{concentration} experiments assess the suitability of set-difference-captioning approaches to sparse, real-world differences.
We restrict our experiments to open-weight models to support reproducibility and ease of deployment.
The proposed benchmark and analysis provide a step towards practical, human-interpretable dataset introspection for autonomous driving datasets.
Our implementation and benchmark dataset are available at \url{https://github.com/KIT-MRT/AD-Diff}
\end{abstract}


	\section{Introduction}
    \balance 
	\label{sec:introduction}

    {\IEEEPARstart{O}{ver} the last decade, autonomous driving research has shifted more and more towards machine learning approaches, relying on deep-learning-based system components  for detection, tracking, and increasingly also for decision-making and trajectory planning.
    Many in the community 
       \parfillskip=0pt\par}
    \par\vspace{\textfloatsep}\noindent
    believe that the field’s future lies in fully end-to-end autonomous driving systems that are completely data-driven, without any engineered heuristics or model-based parts.
    Consequently, data quantity and quality become primary factors in achieving high system reliability and robustness.

    Much effort has been put into answering the question of how to effectively collect, label, and curate~\cite{greer2025language, kim2025automatic} the large and 
    diverse datasets that are required for a data-driven approach to autonomous driving.
    Complete \enquote{data engines} have been proposed to automate large parts of a continuous record-label-retrain-deploy pipeline~\cite{AIDE, datacentricreview}.

    Yet, very little research addresses tools to help developers and stakeholders understand dataset composition -- essential for adequate system performance.
    Without such data introspection tools, decision-making regarding dataset composition and system operation will always rely either on guesswork or on costly, poorly scalable human labor.

    In this paper, we adapt a set-difference-captioning tool for the autonomous driving research community.
    Given two potentially very large image sets, the goal of \textit{set difference captioning} is to generate natural-language descriptions of their differences.
    In realistic deployments, fleets of autonomous vehicles collect data over diverse sensor configurations, locations, and time periods, yielding large volumes of raw data accompanied by metadata (e.g., timestamps, GNSS positions, data source, etc.) and partial semantic labels (e.g., bounding boxes).
    However, analysis based on metadata and available annotations is constrained by the predefined label space and therefore provides only limited insight.

    A tool that generates natural-language difference descriptions between two image sets enables pairwise comparisons of subsets.
    This supports scalable dataset introspection by revealing distribution shifts, gaps, trends, and biases, including subtle differences that are not readily apparent to human inspection.
    Such differences are directly relevant to safety and reliability when deploying and validating autonomous driving systems across domains.
    They can indicate changes in object appearance, infrastructure, or environmental conditions that are not captured by existing labels and, therefore, can affect downstream model behavior.

    Natural language provides a flexible and expressive summary format that complements fixed metadata and label taxonomies.
    Crucially, the method emphasizes \emph{relative} descriptions: comparing a target set against a reference set highlights salient properties and suppresses incidental variation.
    In contrast, unreferenced set descriptions are either trivial (e.g., \enquote{there are cars}) or become unmanageably exhaustive when applied to high-dimensional image data, where the space of potentially describable properties grows rapidly with dataset size and diversity.
    A reference set is therefore essential to determine which properties are meaningful and relevant for decisions, as opposed to expected or irrelevant background variation.

The proposed approach enables more automated data pipelines while preserving a human-in-the-loop interface for safety-relevant decisions in increasingly automated data collection and management systems. Our key contributions are:
\begin{itemize}
    \item \textbf{Object-centric set difference captioning.} Building on prior work on set difference captioning, we adapt the task to the autonomous driving domain by introducing \emph{object-centric} set difference captioning, a formulation tailored to safety-relevant objects and long-tail visual concepts in driving data.
    \item \textbf{Set difference captioning benchmark for autonomous driving.} We curate a benchmark dataset and evaluation protocol for object-centric set difference captioning, termed AD-Diff~Bench, which we will release publicly to support reproducible comparisons.
    \item \textbf{First study of high-sparsity differences.} We study set difference captioning in the \emph{high-sparsity} regime, clarifying its real-world applicability to distribution shifts and safety-relevant dataset diagnostics.
\end{itemize}

    \section{Related Work}
    \label{sec:relatedwork}
Prior work on dataset introspection spans diverse approaches, yet often relies on predefined categories or additional annotations~\cite{wang2020revise}, or targets narrow aspects such as identifying challenging subsets~\cite{eyuboglu2022domino,deon2021spotlight}, limiting generality. In contrast, we are interested in open-ended, human-interpretable descriptions of differences between subsets. We therefore consider set difference captioning, i.e., generating natural language descriptions that characterize how two image sets differ.

     \subsubsection{Vision Language Models}
    To bridge the gap from images to natural language descriptions, we rely on Vision Language Models (VLMs). Modern VLMs couple high-capacity vision encoders with Large Language Models (LLMs), enabling compositional scene reasoning. Early models like Flamingo~\cite{alayrac2022flamingo} and BLIP\nobreakdash-2~\cite{li2023blip2} demonstrated strong few-shot capabilities by combining frozen image encoders with LLMs. LLaVA~\cite{liu2023visual} introduced \textit{visual instruction tuning}, providing VLMs with reasoning capabilities similar to those of LLMs. The emergence of high-performing open-weight VLMs like Qwen3\nobreakdash-VL~\cite{qwen2025qwen3vl} and InternVL\nobreakdash-3.5~\cite{internvl35} now enables practical deployment of VLM-based systems without relying on proprietary APIs, which is particularly relevant for autonomous driving applications where data privacy and offline operation are important considerations.

    \subsubsection{Difference and Change Captioning}
    Prior work on difference captioning~\cite{alayrac2022flamingo,li2023mimicit,li2023otter,yao2022image} and change captioning~\cite{chang2023changes,kim2021viewpoint,park2019robust} has focused on describing differences between pairs of images. However, scaling these methods to thousands of images would require quadratic comparisons and yield fragmented descriptions, which is a crucial limitation for dataset-level analysis. 

    \subsubsection{Set Difference Captioning}
    In the text domain, the D3 framework~\cite{zhong2022describing} introduced a proposer-ranker approach to describe differences between text distributions using LLMs.
    Based on this prior work in the text domain, Dunlap et al.~\cite{dunlap2024describing} formalized the task of describing differences between \textit{sets of images}. They introduced VisDiff, a two-stage algorithm combining a caption-based proposer using BLIP\nobreakdash-2 and GPT\nobreakdash-4 with a CLIP-based ranker. Their work demonstrated the utility of set difference captioning for understanding dataset shifts, comparing model behaviors, and analyzing generative models. GS\nobreakdash-CLIP~\cite{zhu2022gsclip} proposed a related approach using CLIP features to retrieve differences from a predefined text bank, though it is limited to a fixed vocabulary. Our work builds directly on the VisDiff framework but adapts it to the autonomous driving domain using state-of-the-art open-weight models and introduces an object-centric formulation better suited to complex traffic scenes.

    \subsubsection{Domain Shift Datasets}
    To benchmark set difference captioning, we require image-sets with ground-truth difference descriptions. Since even subtle dataset shift can significantly reduce accuracy~\cite{torralba2011unbiased}, many benchmarks have been proposed to evaluate classifier robustness to domain shift~\cite{recht2019imagenet, hendrycks2021many, liang2022metashift, koh2021wilds, SHIFTCVPR2022}. Of these only MetaShift~\cite{liang2022metashift} includes explicit ground-truth annotations for domain-shift and enough image-sets for a meaningful evaluation of set difference captioning. 
    VisDiffBench~\cite{dunlap2024describing} is specifically designed to evaluate set-difference captioning. 
    It consists of 150 image-set pairs with 100 web-scraped images per set, and additionally 37 image-set pairs from ImageNetR~\cite{hendrycks2021many} and ImageNet*~\cite{imagenetstar}. 
    MetaShift and VisDiffBench contain, however, only few images from road environments.
 We fill this gap by introducing AD\nobreakdash-Diff~Bench, the first set-difference captioning benchmark for autonomous driving, consisting solely of domain-relevant images.

    \section{Set Difference Captioning for Autonomous Driving}
    \label{sec:adsetdifferences}

\begin{figure}
  \centering
  \includegraphics[width=\linewidth]{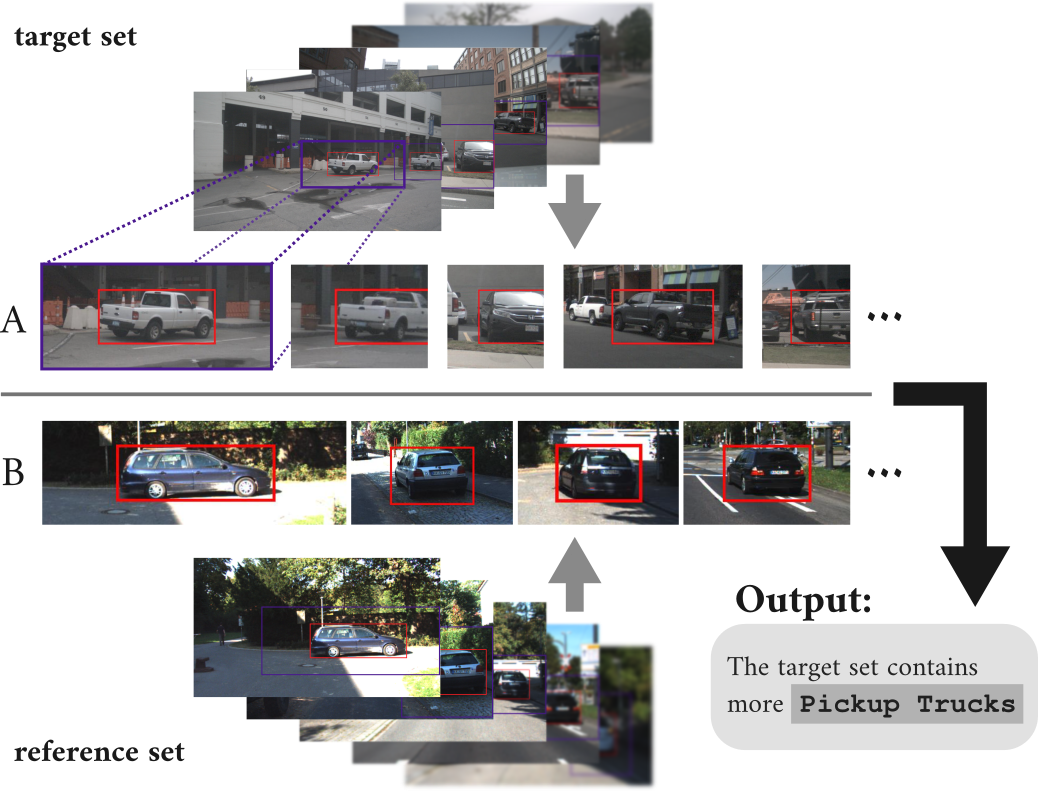}
  \captionof{figure}{\textbf{Object-centric set difference captioning.} Given two image datasets $A$ and $B$, we find difference descriptions between the objects in the two datasets. The output is a natural-language description of a concept that is more true for the \textit{target set} $A$ than for the \textit{reference set} $B$. In this example, $A$ and $B$ are images from the nuImages and KITTI dataset, respectively.}
  \label{fig:task}
\end{figure}

    Our approach to understanding autonomous driving datasets is based on descriptions of the differences between two subsets of the data. The specific question that one wants to answer (e.g., \enquote{How are cars in country \textit{A} different from country \textit{B}?}, \enquote{Which conditions contribute to accidents/close-calls?}) determines the criteria by which the data is split into two subsets (\textit{location} and \textit{accident} label, respectively, for the examples). Once the subsets are defined, the task is reduced to a \textit{set difference captioning} task~\cite{dunlap2024describing}. In this section, we first define the task of set difference captioning before aligning this more general task definition to the specific requirements in the autonomous driving domain, introducing the task of \textit{object-centric set difference captioning}.
    \subsection{Set Difference Captioning}
    \label{sec:taskdefinition}
    The goal of the set difference captioning task (Fig. \ref{fig:task}) is to generate natural-language descriptions of the differences between two image sets \textit{A} and \textit{B}. 
    We follow prior work~\cite{dunlap2024describing} by defining a correct difference description as a description that is more true for set \textit{A} than for set \textit{B}. 
    We refer to set \textit{A} as the \textit{target set} and to set \textit{B} as the \textit{reference set}. The ground truth difference descriptions in our benchmark are therefore descriptions of the target set \textit{A}. A difference description is considered correct if it is semantically equivalent to the ground truth difference description.

    \subsection{Object-centric Set Difference Captioning}
    \label{sec:ocsetdiffcaptioning}
    Directly applying set difference captioning to camera images from autonomous driving datasets will, in most cases, not result in useful difference descriptions. Any single image typically contains many traffic participants and other objects; difference descriptions between complete images are therefore prone to missing relevant differences. Furthermore, descriptions would accumulate information about different objects in one \textit{or} multiple images in natural language (e.g., \enquote{3 more people with red shirts}), making automated aggregation across sets difficult.

    We therefore introduce a modification of the original task that we term \textit{object-centric set difference captioning}. The goal of this task is to caption the difference between two sets of objects observed in the sensor data. Generally, but not necessarily, the objects in both sets are of the same class, as differences between objects of different classes are rarely informative -- there is no point in \enquote{comparing apples to oranges}. This object-centric approach resolves aggregation issues and makes it straightforward to ensure that no objects are ignored during difference captioning. If required, it is also easier with this approach to attribute difference descriptions to specific object instances.

    To adapt existing set-difference captioning methods to the object-centric setting, we first select one image in which the object appears, if it is tracked across multiple time steps or across cameras. We then extract an image patch centered on the object from the corresponding camera image, as shown in Fig.~\ref{fig:task}. The object is localized using pre-trained 2D detection models or bounding box annotations, if available. These image patches can be partitioned into subsets based on dataset annotations or metadata and then compared using the same procedure as in general set-difference captioning. Some differences between sets are only identifiable given additional context from the image. Therefore, we enlarge the image patches by \SI{50}{\percent} from the raw bounding box annotations. The object of interest is marked with a red bounding box painted on the image patch, allowing the set-difference-captioning approach to locate the correct object even in crowded scenes.

    \section{Benchmark}
    \label{sec:benchmark}

    The only existing benchmark for set difference captioning of images~\cite{dunlap2024describing} targets general-purpose use. The domain shift to camera images recorded from autonomous vehicles is therefore high. We introduce AD\nobreakdash-Diff~Bench, a benchmark for \textit{object-centric set difference captioning}, consisting of images from the autonomous driving domain and taking requirements for real-world use in this application into account. This section first introduces the datasets we collected for AD\nobreakdash-Diff~Bench before explaining techniques to test whether set captioning approaches also work on sparse or noisy sets, which are common in the real world.

    \subsection{Set Differences Dataset}
    \label{sec:dataset}
\begin{figure}
    \centering
    \includegraphics[width=1\linewidth]{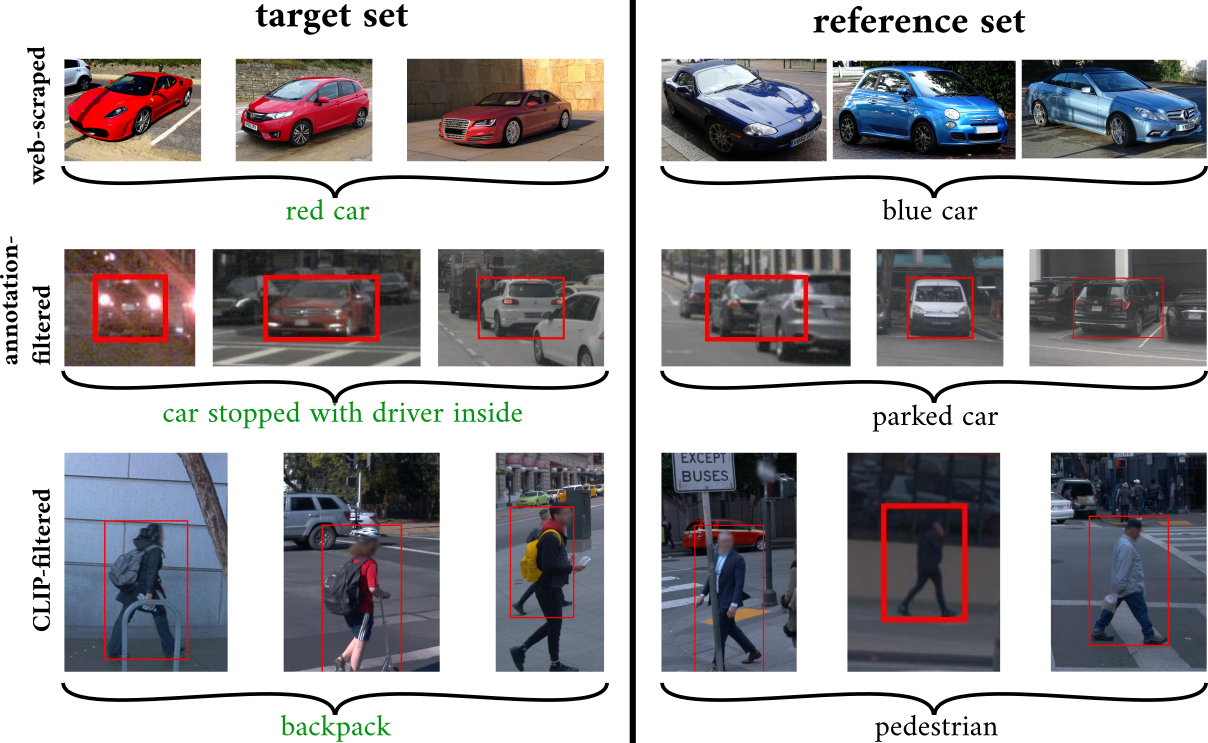}
    \caption{\textbf{Splits of AD-Diff Bench.} Example set-pairs for web-scraped, annotation-filtered, and CLIP-filtered split. Images in the annotation-filtered and CLIP-filtered split are extracted from KITTI, nuImages, and Waymo Open Dataset. Each row shows three images from each set together with the ground-truth set description. The ground-truth set-pair difference description is the description of the target set, marked in green.}
    \label{fig:examples}
\end{figure}

    \begin{table}[]
    \centering
    \caption{Number of sets and images in each split of AD-Diff Bench.}
    \begin{tabular}{l c c}
    \textbf{Dataset} & \textbf{\# Paired Sets} & \textbf{\# Images Per Set} \\
    \hline
        Web-scraped & 60+60+60 & 100 \\
        Annotation-filtered & 60 & 10-8000\\
        CLIP-filtered & 80 & 100
    \end{tabular}
    \label{tab:sumarrybench}
    \end{table}

    To compare set-difference-captioning approaches with respect to their applicability as a development tool for real autonomous driving systems, a sufficiently large dataset of domain-relevant image set pairs with ground truth difference descriptions must be collected first. We consider images of traffic scenes, road users, and any object that could reasonably appear on public roads as domain-relevant. To make the benchmark challenging enough, all images of an image set pair display traffic-relevant objects of the same object category, with the difference between the two image sets being subtle enough to be interesting in the context of a set difference captioning task.
    
    At the same time, differences across the dataset should be as diverse as possible, as set captioning approaches should be applicable to any possible semantic difference between image set pairs. To this end, we first manually craft a list of set-differences we want to include in the dataset, with each set difference being defined by target-captions or filter criteria for the target image set ($A$) and the reference image set ($B$) of each image set pair. Based on this, we use three different approaches to collect the images for the image sets, resulting in the three datasets of image set pairs with ground truth difference annotations that make up AD\nobreakdash-Diff~Bench. Table~\ref{tab:sumarrybench} summarizes the three datasets of our benchmark, Fig.~\ref{fig:examples} shows example images for each split.

    \subsubsection{Web-scraped}
    \label{sec:webscraped}
 Similarly to VisDiffBench~\cite{dunlap2024describing}, a benchmark for general image set difference captioning, our first dataset is collected by querying Bing image search for the pre-defined target-captions. Each image set consists of the top 100 results retrieved from Bing image search for its target-caption. The ground-truth difference description of the image set pair is then the search query of the target image set ($A$). We manually filter out image set pairs that do not display the intended target-difference or that prominently display differences other than the ground-truth difference. We collect 180 image set pairs, grouped by expert knowledge into three difficulty levels and 6 domain-relevant object categories: pedestrians, cars and light trucks, micromobility (bicycles, e-scooters, etc.), large vehicles, traffic control devices, and misc; with 10 image set pairs per object category and difficulty level. The resulting dataset is very diverse in both images and difference-descriptions, including both common and rare (corner-case) representations of road users and traffic-relevant objects.

    \subsubsection{Annotation-filtered}
    \label{sec:annotationfiltered}
    The web-scraped dataset is very diverse but shows a significant domain shift away from images recorded by autonomous vehicles (camera system, perspective, sampling bias). To be able to draw confident conclusions regarding the applicability of set-difference-captioning approaches to data from the autonomous driving domain, we additionally create a dataset of image set pairs sampled from three different autonomous driving datasets: KITTI~\cite{geiger2013kitti}, nuImages~\cite{nuImages}, and Waymo Open Perception~\cite{sun2020waymo,waymo2}.
    
    In line with the object-centric approach, both image sets of each pair consist of image patches featuring objects of the same class, cut from the bounding box annotations of the three autonomous driving datasets. The filter criteria that determine which images are assigned to which image set are defined based on the fine-grained annotations for sub-classes and attributes contained in these datasets, resulting in very clean and well-defined image sets. Difference descriptions are manually defined from the filtered attributes. 
    
    The image set pairs of this dataset vary in image set size from 10 to \num{8000} images, testing not only the accuracy but also the scalability of set-difference-captioning approaches. We provide two sub-splits of this dataset: the main split annotation\nobreakdash-filtered\nobreakdash-60, including all 60 image set pairs of this split, and the subset annotation\nobreakdash-filtered\nobreakdash-39, which is primarily intended for experiments with the concentration parameter that we will introduce in Section \ref{sec:dilution}.
    
    \subsubsection{CLIP-filtered}
    \label{sec:clipfiltered}
    While curating image sets based on already available annotations is convenient and results in very accurate and clean set-pairs, diversity is limited by the predefined categories and attributes. To obtain more set-pairs with low domain shift from data collected by deployed autonomous vehicles, we reuse the image patches extracted from the autonomous driving datasets for the annotation-filtered dataset, but use CLIP instead of annotations to filter the image patches into subsets. For each image set, we use OpenClip~\cite{openclip} to filter all image patches of the respective object-category for the 100 best matching image patches given the predefined keyword (highest cosine similarity between the image-embedding of the image patch and the text-embedding of the keyword). Keywords are based on both descriptions of the target set and the reference set of the web-scraped dataset. Sets containing too many incorrect matches are discarded by manual inspection. The 80 pairs of image sets of this dataset consist of the keyword ground-truth description, the CLIP-filtered image patches for the target set, and 100 image patches randomly sampled from the same category for the reference set.

    \subsection{Set Difference Captioning of Diluted Sets}
    \label{sec:dilution}
    
    The image sets in AD\nobreakdash-Diff~Bench are relatively clean, i.e.,  every image set contains primarily images that match the corresponding set description. Sets of images collected in the real world are, however, rarely clean and usually contain images of objects with many different appearances that do not all fit a singular set description. 
    
    The VisDiffBench set difference captioning benchmark~\cite{dunlap2024describing} introduced the concept of \textit{purity} to account for this mismatch between reality and the benchmark. Purity is a parameter that can be set for each run of the benchmark to a value between 0 and 1 to test the performance of set captioning approaches on noisy sets, i.e.,  sets with partially incorrect labels. A purity value of 1 indicates a completely clean, i.e., unchanged set pair. Lower values indicate that some portion of the images has been swapped between the two sets, with half the images being swapped for a value of zero, leading to completely shuffled sets. A set-difference-captioning approach that works even for low purity values would, for example, be able to detect, given sets of images of pedestrians from two different cities, that pedestrians wear hats more often in one of the cities. A set-difference-captioning approach that works only for purity values close to one will only detect such a difference if hats are only ever worn in one of the two cities.

    Sparse differences between sets are equally relevant and common. The difference between two sets is sparse if the correct difference description is only true for a small fraction of the images in each set. If there were, for example, a new fad for pink-colored cars, which were previously nearly non-existent, then the difference observed between an image set recorded after the trend has been established and an image set recorded years earlier would be a sparse difference. Not all cars are colored pink with the new trend; non-pink cars are still much more common. Images fitting the difference description appear in this case only in one of the sets but are rare even in this set.
    This is particularly important for the development of safe and reliable autonomous driving systems, which requires identifying even very rare potential hazards before they become relevant.

\begin{figure*}
    \centering
    \includegraphics[width=0.8\linewidth]{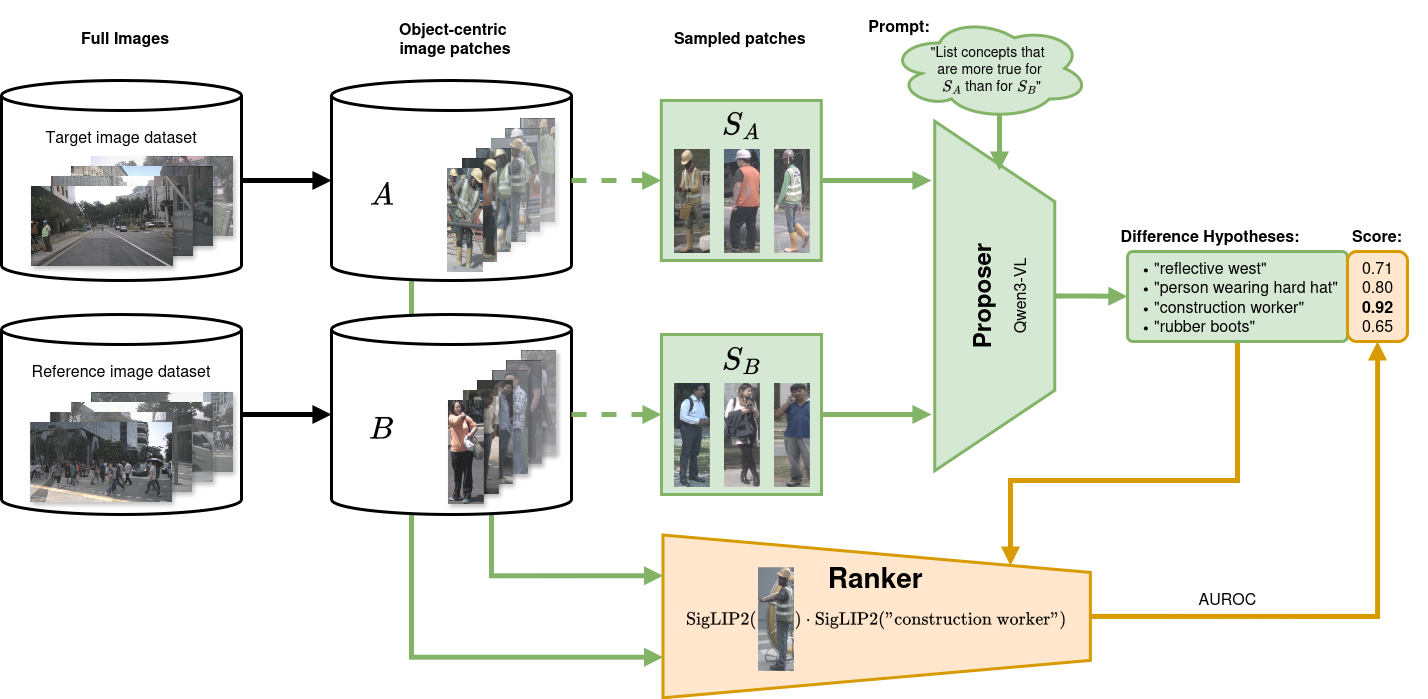}
    \caption{\textbf{Object-centric set difference captioning} using the two-stage, image-based approach. First, object-centric image patches are extracted from the target image set and reference image set, resulting in the target and reference patch sets $A$ and $B$, respectively. Then, the \textit{proposer} predicts difference hypotheses based on subsets of image patches $S_A$ and $S_B$ sampled from $A$ and $B$. Lastly, the \textit{ranker} evaluates the difference hypotheses from the \textit{proposer} on the complete datasets $A$ and $B$, calculating a score for each hypothesis. Adapted from~\cite{dunlap2024describing}}
    \label{fig:approach}
\end{figure*}

    To test the capability of set-difference-captioning approaches to detect sparse differences, we introduce the concept of dilution to AD\nobreakdash-Diff~Bench. A set pair is diluted by adding additional images from a third set to both sets. Dilution is measured by the parameter concentration; the more diluted the set, the lower the concentration. An image set $S$ of $n_S$ images is diluted to a concentration of $c=\frac{n_S}{n_S + n_D}$, with $n_D$ being the number of dilution images sampled from the dilution set $D$ and added to the set $S$. Each set pair in the annotation-filtered split of AD\nobreakdash-Diff~Bench is assigned a  dilution set that is selected to be similar enough to the target and reference set to make high dilution experiments challenging. For some set-pairs, the dilution set can overlap or be identical to the reference set, but there is never any overlap between the target set and the dilution set.

    We consider the accurate detection of even very sparse differences to be of paramount importance for a set difference captioning tool when used for autonomous driving applications. To illustrate this, consider that out of the \num{784444} annotations for the class \textit{vehicle} in the nuImages~\cite{nuImages} dataset, only 42 correspond to the sub-class \textit{ambulance}. This corresponds to a concentration of $c\approx \num[scientific-notation=true]{0.000054}$. This value is tiny, but correctly identifying ambulances is without a doubt still important for autonomous vehicles.
    \subsection{Evaluation}
    \label{sec:evaluation}
    To evaluate whether the generated difference captions should be considered correct, we employ the open-weight LLM gpt\nobreakdash-oss\nobreakdash-120b~\cite{openai2025gptoss120bgptoss20bmodel} as a judge. gpt\nobreakdash-oss is tasked with evaluating whether the ground-truth difference description from the dataset is semantically equivalent to the difference description proposed by the evaluated approach. The prompt given to gpt\nobreakdash-oss includes exact scoring rules for edge cases where the decision between clearly semantically equivalent and not equivalent is not obvious. Based on these scoring rules, gpt\nobreakdash-oss can give a score of 0 (no match), 0.5 (partial match), and 1 (perfect match). The highest score among the top-N ranked hypotheses for each set pair is averaged across all set pairs to compute the top-N accuracy (Acc@N).

    To evaluate whether the assessment of gpt\nobreakdash-oss is reliable, we manually labeled over \num{1000} hypotheses on VisDiffBench~\cite{dunlap2024describing} using the same scoring rules that we provide to gpt\nobreakdash-oss in the prompt. We find that gpt\nobreakdash-oss assigns the same score as the human annotator for \SI{80.3}{\percent} of the hypotheses, with a low mean absolute error of \num{0.104}.

    \section{Approach}
    \label{sec:approach}

    Our general approach follows~\cite{dunlap2024describing} with the distinction between separate \emph{proposer} and \emph{ranker} submodules, illustrated in Fig.~\ref{fig:approach}.
    We consider two datasets, the target set $A$ and the reference set $B$, for which we seek additional insights.
    Since comparing large datasets directly using vision-language models is still not feasible, the proposer takes a subset ${S_A \subseteq A}, {S_B \subseteq B}$ of the two input datasets and generates a list of difference hypotheses.
    This is followed by the ranker, which compares these hypotheses against the full datasets $A, B$ and assigns a score to each hypothesis.
    The final explanation for perceived differences between the two datasets is then the top ranked hypothesis.
    For transparency and reproducibility, we use publicly available, open-weight state-of-the-art models.

    \subsection{Proposer}
    We employ one of three methods to generate difference hypotheses: image-based, caption-based, or feature-based.
    To ensure a fair comparison, all three proposer methods use Qwen3-VL-30B-A3B-Instruct as the VLM and LLM.
    We perform three generation rounds.
    In each round, we independently sample 20 images from each set $A$ and $B$ and generate ten difference hypotheses.

    \subsubsection{Image-based}
    Since modern VLMs accept multiple images as input, we just feed the subsets $S_A, S_B$ directly into the model together with the prompt.

    \subsubsection{Caption-based}
    In a first step, we generate captions for all images in the sampled subsets $S_A, S_B$ with the help of the VLM.
    We then compile the captions into a new text-only prompt and reprocess them with the LLM to generate the hypothesis.  
    
    \subsubsection{Feature-based}
    We start by extracting visual embeddings from the VLM for all images in $S_A$ and $S_B$, and compute the mean embedding for each subset.
    Using feature arithmetic, we take the difference between the mean embeddings and use it as a direction that captures the shift from $A$ to $B$.
    We combine this difference representation with prompt text embeddings and pass the result through the VLM decoder to produce the hypothesis.
    To enable averaging, all embeddings must have the same shape, which we achieve by scaling all images to a size of $224 \times 224$.

        \begin{table*}
        \centering
        \caption{Results of methods on the three splits of AD-Diff Bench, averaged over three runs.}
        \begin{tabular}{c c|c c|c c|c c}
        \multirow[c]{2}{4em}{\textbf{Approach}} & \multirow[c]{2}{4em}{\textbf{Proposer}} & \multicolumn{2}{c}{\textbf{Web-scraped}} & \multicolumn{2}{c}{\textbf{Annotation-filtered-60}} & \multicolumn{2}{c}{\textbf{CLIP-filtered}} \\
        & & Acc@1 & Acc@5 & Acc@1 & Acc@5 & Acc@1 & Acc@5 \\
        \hline
            \multirow{3}{5em}{Two-stage}  & Image-based & \textbf{0.73} & \textbf{0.88} & 0.56 & 0.78 & \textbf{0.64} & \textbf{0.80} \\
              & Caption-based & 0.70 & 0.83 & \textbf{0.60} & \textbf{0.83} & \textbf{0.63} & \textbf{0.81} \\
              & Feature-based & 0.33 & 0.45 & 0.20 & 0.28 & 0.41 & 0.55 \\
            \hline
            Single-stage & Image-based & 0.64 & 0.85 & 0.53 & 0.70 & 0.49 & 0.62 \\
        \end{tabular}
        \label{tab:results}
    \end{table*}

    \subsection{Ranker}
    Since the hypotheses generated from the proposer have high variance due to the subsampling of data, a second filtering step is required.
    To this end,~\cite{dunlap2024describing} introduced three ranking methods.
    We focus on the feature-based ranker as it substantially outperformed the alternatives and is the only method that produces continuous values.
    Given hypothesis $h$, this ranker computes $R(x, h)$ for all images $x \in A \cup B$, where $R(x, h)$ is the cosine similarity between the image embedding of $x$ and the text embedding of $h$.
    Interpreting $R$ as a binary classifier score, the resulting AUROC is used as the final score for the hypothesis.
    We perform ranking using a state-of-the-art SigLIP 2 Giant model~\cite{siglip2}.
    
    \subsection{Single-stage}
    In addition to the two-stage pipeline, we propose a single-stage variant that unifies the proposer and ranker in a single inference step.
    This is feasible because modern VLMs can jointly process a substantially larger number of input images.
    After subsampling 100 images per dataset, they are given to the VLM together with a prompt detailing the task to find differences among the datasets, as well as ranking them.
    With this setup, the VLM generates difference hypotheses directly in one pass.

    \section{Results}

    We present experimental results addressing four questions: 
    (A)~How do the three dataset splits compare?
    (B)~Is a two-stage approach necessary for reliable set difference captioning?
    (C)~Which proposer performs best?
    (D)~Can current methods handle real-world sparse and noisy set differences?
    (E)~Can we use object-centric set difference captioning to get new insights about dataset composition? 

    \subsection{Comparison between Dataset Splits}
    As can be seen in Table~\ref{tab:results}, accuracy scores are lowest for most approaches on the annotation-filtered split and highest on the web-scraped split. The image patches in the two splits based on autonomous driving datasets, annotation-filtered and CLIP-filtered splits, are generally of much lower resolution than those of the web-scraped split and are often further degraded by image noise, motion blur, or underexposure, making these two splits inherently more difficult than the web-scraped split. It is also possible that the VLMs used in our experiments perform worse on autonomous driving data, regardless of resolution or image quality, due to the larger domain shift from their (primarily web-scraped) training data. Regardless of which factor contributes more to the observed performance difference, the fact that data from autonomous driving datasets is generally more challenging for set-difference-captioning approaches highlights the need for a dedicated, domain-specific benchmark.
    
    The annotation-filtered split is harder than the CLIP-filtered split not due to the images, but due to the differences described. A significant amount of differences in the annotation-filtered split are hard to recognize even for humans (e.g., \enquote{parked car} vs \enquote{car stopped with driver inside}), while the CLIP-filtered split features mostly differences that significantly affect appearance. The sets of the annotation-filtered split are, however, much cleaner than in the web-scraped and CLIP-filtered sets.

    We furthermore validate the difficulty levels of the web-based split (easy, medium, hard) by comparing the accuracy of the image-based two-stage approach on these sub-splits. As expected, the acc@1 is highest for the easy split and lowest for the hard split ($0.84 > 0.75 > 0.59$).

    \subsection{Two-stage or Single-stage Approach?}
    Comparing the two-stage image-based approach to the single-stage image-based approach in Table~\ref{tab:results}, the two-stage approach results in much more accurate difference captions for the web-scraped and CLIP-filtered splits, but it performs similarly to the single-stage approach on the annotation-filtered split. We attribute this to a greater robustness of the two-stage approach with respect to higher sparsity of the target and reference sets. The feature-based ranker can effectively aggregate even very sparse differences across the complete dataset, which the single-stage approach cannot correctly rank from the small samples it sees. 
    
    This hypothesis is supported by the dilution experiments described in Section \ref{sec:resdil}. The web-scraped and CLIP-filtered datasets are sparser than the annotation-filtered split, containing images that do not fit either the target or the reference set description due to the inherent limitations of search engines and CLIP-based similarity search, compared to manual annotations. In this case, the single-stage approach is significantly outperformed by the two-stage approach, but not for the very clean annotations of the annotation-filtered split. In general, the two-stage approach seems to be the more robust choice for sparse datasets from the real world. The single-stage approach might be a good choice if the goal is to quickly sample difference hypotheses, as it can be orders of magnitude faster for large datasets.

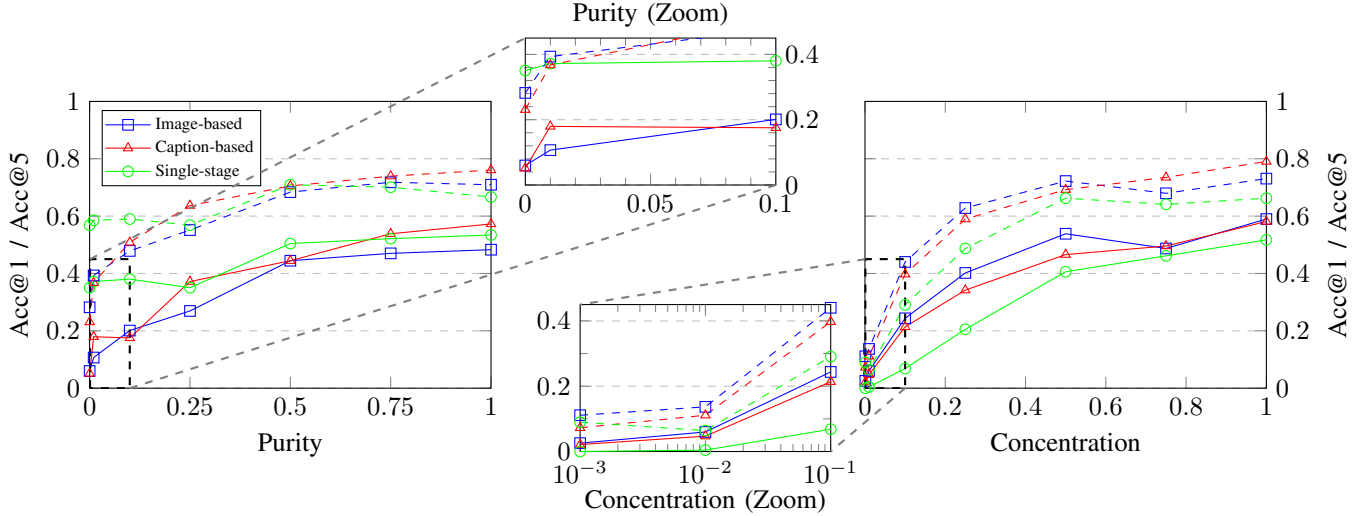
\begin{figure*}[]
\centering
\begin{tikzpicture}

\pgfmathsetlengthmacro{\FullW}{0.38*\textwidth}
\pgfmathsetlengthmacro{\ZoomW}{0.27*\textwidth}
\pgfmathsetlengthmacro{\FullH}{0.78*\FullW}
\pgfmathsetlengthmacro{\ZoomH}{0.72*\ZoomW}
\pgfmathsetlengthmacro{\XsepMT}{0.025*\textwidth}
\pgfmathsetlengthmacro{\XsepMB}{0.065*\textwidth}
\pgfmathsetlengthmacro{\XsepR}{0.025*\textwidth}
\pgfmathsetlengthmacro{\Ysep}{0.0*\textwidth}

\pgfmathsetlengthmacro{\ZoomShift}{0.5*(\ZoomH+\Ysep)}

\begin{axis}[
    name=purityFull,
    width=\FullW,
    height=\FullH,
    xlabel={Purity},
    ylabel={Acc@1 / Acc@5},
    ymajorgrids=true,
    grid style=dashed,
    legend pos=north west,
    xmin=0, xmax=1,
    ymin=0, ymax=1,
    xtick={0,0.25,0.5,0.75,1},
    legend style={
        font=\footnotesize,
        nodes={scale=0.8, transform shape},
        inner sep=1pt,
        legend cell align={left},
        legend columns=1,
    },
]
\addplot[color=blue, mark=square]
    table [x=purity, y=acc1, col sep=comma]
    {data/purity_VLM.csv};
\addlegendentry{Image-based}

\addplot[color=red, mark=triangle, mark options={solid}]
    table [x=purity, y=acc1, col sep=comma]
    {data/purity_LLM.csv};
\addlegendentry{Caption-based}

\addplot[color=green, mark=o]
    table [x=purity, y=acc1, col sep=comma]
    {data/purity_zero.csv};
\addlegendentry{Single-stage}

\addplot[color=blue, mark=square, mark options={solid}, dashed]
    table [x=purity, y=acc5, col sep=comma]
    {data/purity_VLM.csv};
\addplot[color=red, mark=triangle, mark options={solid}, dashed]
    table [x=purity, y=acc5, col sep=comma]
    {data/purity_LLM.csv};
\addplot[color=green, mark=o, mark options={solid}, dashed]
    table [x=purity, y=acc5, col sep=comma]
    {data/purity_zero.csv};

\draw[black, thick, dashed]
    (axis cs:0.001,0)
    rectangle
    (axis cs:0.1,0.45);

\coordinate (purityRectNW) at (axis cs:0.001,0.45);
\coordinate (purityRectSE) at (axis cs:0.1,0);
\end{axis}

\begin{axis}[
    name=purityZoom,
    at={(purityFull.east)},
    anchor=west,
    xshift=\XsepMT,
    yshift=\ZoomShift,
    width=\ZoomW,
    height=\ZoomH,
    xlabel={Purity (Zoom)},
    xlabel style={at={(axis description cs:0.5,1.3)},anchor=north},
    ymajorgrids=true,
    yticklabel pos=right,
    grid style=dashed,
    xmin=0.0, xmax=0.1,
    ymin=0, ymax=0.45,
    xtick={0,0.05,0.1},
    minor tick num=4,
    xticklabel style={
        /pgf/number format/precision=2,
        /pgf/number format/fixed,
    },
]
\addplot[color=blue, mark=square]
    table [x=purity, y=acc1, col sep=comma]
    {data/purity_VLM.csv};
\addplot[color=red, mark=triangle]
    table [x=purity, y=acc1, col sep=comma]
    {data/purity_LLM.csv};
\addplot[color=green, mark=o]
    table [x=purity, y=acc1, col sep=comma]
    {data/purity_zero.csv};
\addplot[color=blue, mark=square, mark options={solid}, dashed]
    table [x=purity, y=acc5, col sep=comma]
    {data/purity_VLM.csv};
\addplot[color=red, mark=triangle, mark options={solid}, dashed]
    table [x=purity, y=acc5, col sep=comma]
    {data/purity_LLM.csv};
\addplot[color=green, mark=o, mark options={solid}, dashed]
    table [x=purity, y=acc5, col sep=comma]
    {data/purity_zero.csv};
\end{axis}

\begin{axis}[
    name=concentrationZoom,
    at={(purityFull.east)},
    anchor=west,
    xshift=\XsepMB,
    yshift=-\ZoomShift,
    width=\ZoomW,
    height=\ZoomH,
    xlabel={Concentration (Zoom)},
    xlabel style={at={(axis description cs:0.5,-0.2)},anchor=north},
    ymajorgrids=true,
    grid style=dashed,
    xmode=log,
    xmin=0.001, xmax=0.1,
    ymin=0, ymax=0.45,
    minor tick num=1,
]
\addplot[color=blue, mark=square]
    table [x=concentration, y=acc1, col sep=comma]
    {data/concentration_VLM.csv};
\addplot[color=red, mark=triangle]
    table [x=concentration, y=acc1, col sep=comma]
    {data/concentration_LLM.csv};
\addplot[color=green, mark=o]
    table [x=concentration, y=acc1, col sep=comma]
    {data/concentration_zero.csv};
\addplot[color=blue, mark=square, mark options={solid}, dashed]
    table [x=concentration, y=acc5, col sep=comma]
    {data/concentration_VLM.csv};
\addplot[color=red, mark=triangle, mark options={solid}, dashed]
    table [x=concentration, y=acc5, col sep=comma]
    {data/concentration_LLM.csv};
\addplot[color=green, mark=o, mark options={solid}, dashed]
    table [x=concentration, y=acc5, col sep=comma]
    {data/concentration_zero.csv};
\end{axis}

\begin{axis}[
    name=concentrationFull,
    at={(concentrationZoom.east)},
    anchor=west,
    xshift=\XsepR,
    yshift=\ZoomShift,
    width=\FullW,
    height=\FullH,
    xlabel={Concentration},
    ylabel={Acc@1 / Acc@5},
    yticklabel pos=right,
    ymajorgrids=true,
    grid style=dashed,
    xmin=0, xmax=1,
    ymin=0, ymax=1,
]
\addplot[color=blue, mark=square]
    table [x=concentration, y=acc1, col sep=comma]
    {data/concentration_VLM.csv};
\addplot[color=red, mark=triangle]
    table [x=concentration, y=acc1, col sep=comma]
    {data/concentration_LLM.csv};
\addplot[color=green, mark=o]
    table [x=concentration, y=acc1, col sep=comma]
    {data/concentration_zero.csv};

\addplot[color=blue, mark=square, mark options={solid}, dashed]
    table [x=concentration, y=acc5, col sep=comma]
    {data/concentration_VLM.csv};
\addplot[color=red, mark=triangle, mark options={solid}, dashed]
    table [x=concentration, y=acc5, col sep=comma]
    {data/concentration_LLM.csv};
\addplot[color=green, mark=o, mark options={solid}, dashed]
    table [x=concentration, y=acc5, col sep=comma]
    {data/concentration_zero.csv};

\draw[black, thick, dashed]
    (axis cs:0.001,0)
    rectangle
    (axis cs:0.1,0.45);

\coordinate (concentrationRectNW) at (axis cs:0.001,0.45);
\coordinate (concentrationRectSE) at (axis cs:0.1,0);
\coordinate (concentrationRectSW) at (axis cs:0.001,0);
\end{axis}

\draw[thick, gray, dashed] (purityRectNW) -- (purityZoom.north west);
\draw[thick, gray, dashed] (purityRectSE) -- (purityZoom.south east);

\draw[thick, gray, dashed] (concentrationRectNW) -- (concentrationZoom.north west);
\draw[thick, gray, dashed] (concentrationRectSE) -- (concentrationZoom.south east);

\end{tikzpicture}

\caption{\textbf{Results for noisy and diluted sets.} Results of purity and concentration experiments averaged over three runs on the annotation-filtered-39 split.
Solid lines show top-1 accuracy (Acc@1) and dashed lines top-5 accuracy (Acc@5). Full-view plots are shown left (purity) and right (concentration), with zoomed-in plots for very low purity (top) and concentration (bottom) values stacked in the middle.}
\label{fig:purity_concentration_results}
\end{figure*}

    \subsection{Comparison between different Proposers}
    Comparing the different proposers in Table~\ref{tab:results}, the feature-based proposer performs significantly worse than both the image-based and caption-based proposers. This is in line with previous results~\cite{dunlap2024describing}, where the feature-based proposer was able to compete with the alternatives for style differences, but not for semantic differences. Unlike the results of the VisDiff~\cite{dunlap2024describing} approach on VisDiffBench, we did not observe a significant advantage of the caption-based approach over the image-based approach. The accuracy of these two proposers is about the same across all three splits of our dataset. We attribute this to the recent advancements in VLMs.
    Unlike VisDiff's single-image VLM -- requiring all images to be downscaled and composited into one -- modern VLMs accept multiple images natively, making the image-based approach now the simpler and marginally more efficient choice.
    
    \subsection{Noisy and Diluted Sets}
    \label{sec:resdil}

    Fig. \ref{fig:purity_concentration_results} shows the results of our dilution experiments on the annotation\nobreakdash-filtered\nobreakdash-39 split of AD\nobreakdash-Diff~Bench. This split is a sub-split of the annotation\nobreakdash-filtered\nobreakdash-60 split, filtered to include only the image-set pairs with a dilution set large enough to achieve a concentration value below \num{0.001}. 

    As expected, lower concentration values result in lower accuracy scores for all approaches. Interestingly, the accuracy remains stable until a concentration of about \num{0.5} (every second image is from the dilution set), before it drops rapidly for smaller concentrations. None of the tested approaches is able to reliably describe set differences for very low concentrations. 
    
    Accuracy drops much more rapidly for the single-stage approach compared to two-stage approaches. This indicates that the ranker of the two-stage approach can successfully aggregate even very sparse differences across the complete dataset, which the single-stage approach cannot correctly rank from the small sample-size it sees.

    Purity has a similar influence on accuracy as concentration does, with accuracy being fairly stable up to a value of \num{0.5}, but then approaching zero for low purity values. The exception is the single-stage proposer, which performs very poorly for low concentration values but performs well for low purity values. Even at a purity value of \num{0} it is still able to correctly predict the set difference in \SI{35}{\percent} of cases. This is surprising, as the images of both sets are completely shuffled for purity \num{0}, with the two sets being statistically identical. However, the two original sets can still be identifiable to the proposer as clusters of specific image representations within the shuffled set. These differences are averaged out in the ranker, which averages similarity scores between hypothesis descriptions and images across each subset. The ranker is therefore not able to robustly rank the "correct" hypothesis highest if both image sets are on average identical. However, the fact that the one-stage approach can retrieve the original hypothesis should not be considered an advantage, as such a difference description would be considered a false positive in any real scenario. Two image sets featuring distinct clusters of representations within the set, but no statistical differences between the sets, are identical sets from the perspective of set difference captioning.

    This also highlights the advantage of accuracy for low concentration over accuracy for low purity as a metric for the real-world performance of set-difference-captioning approaches, as our results seem to indicate that it cannot be cheated as easily as low purity experiments.

    \subsection{Application Example}

        \begin{figure*}
    \footnotesize
        \centering
        \includegraphics[width=\linewidth]{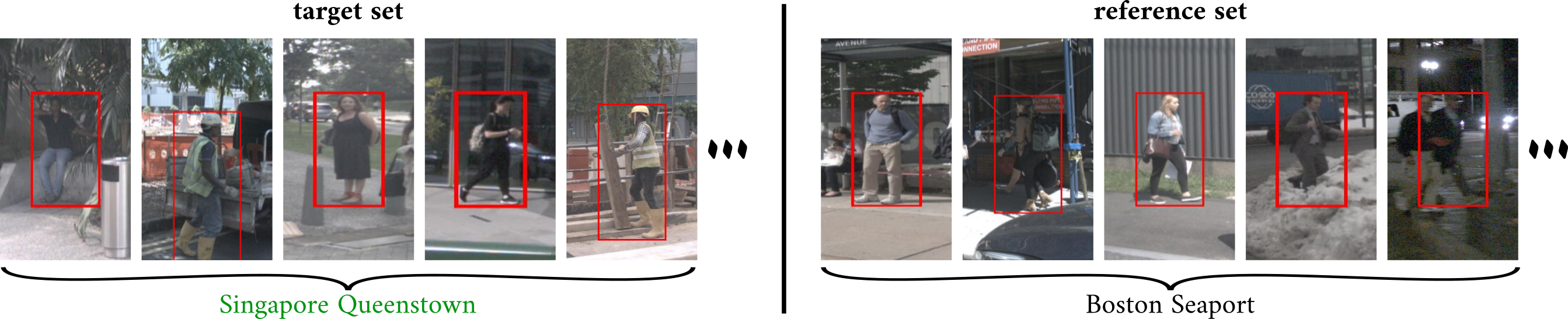}
        
        \vspace{1.4em}
        
        \begin{tabular}{c l c p{12cm}}
            \textbf{\#} & \textbf{Difference Hypothesis} & \textbf{AUROC}\rlap{ $\downarrow$} & \textbf{Supporting Evidence}  \\
            1 &  people wearing hard hats & \num{0.602}  & Fraction of persons labeled \texttt{construction\_worker} in nuImages: Singapore~\SI{12.4}{\percent}, Boston~\SI{3.4}{\percent}\newline Govt. statistics, employed in construction: Singapore~\SI{12.1}{\percent}~(2019)~\cite{singaporeconstr}, Boston~\SI{2.6}{\percent}~(2017)~\cite{bostonconstr} \\
2 & outdoor scenes with greenery & \num{0.585}  & Large parks (Kent Ridge Park, HortPark, etc.) in Singapore Queenstown; no large parks in Boston Seaport \\
3 & construction workers  & \num{0.584}  & \textit{See hypothesis \#1} \\
4 & people in daylight conditions   & \num{0.574}  & Images recorded 8pm -- 6am according to nuImages annotations: Singapore \SI{0.6}{\percent}, Boston \SI{3.4}{\percent} \\
5 & people in high-visibility vests  & \num{0.551}  & \textit{See hypothesis \#1} \\
6 & people wearing work vests   & \num{0.547}  & \textit{See hypothesis \#1} \\
7 & people wearing t-shirts & \num{0.542}  & Average yearly temperature: Singapore~\SI{27.8}{\celsius}~\cite{climatesingapore}, Boston~\SI{11.0}{\celsius}~\cite{climateboston} \\
8 & people in construction zones   & \num{0.541}  & \textit{See hypothesis \#1} \\
        \end{tabular}
        \caption{\textbf{Application example}. Geographic variation in pedestrian appearance and visual context. Comparison of pedestrian image sets from two cities using the proposed two-stage approach with an image-based proposer. The target set contains pedestrian patches from Singapore Queenstown, while the reference set contains pedestrian patches from Boston Seaport. Object-centric image patches were extracted from the nuImages dataset and assigned to each set using location annotations. The generated difference hypotheses are ranked by the AUROC score computed by the ranker. For each hypothesis, we additionally provide supporting evidence based on both nuImages annotations and external sources such as public statistics to assess its plausibility and correctness.}
        \label{fig:qualitative}
    \end{figure*}   
    
    Fig.~\ref{fig:qualitative} shows an application example in which we compare pedestrian image patches from two different cities, Singapore Queenstown and Boston Seaport. One possible application scenario is the expansion of a vehicle fleet's operational design domain (ODD) from Boston to Singapore. The image patches are extracted from nuImages pedestrian bounding box annotations and assigned to their respective set using the location metadata available in nuImages. 
    
    We use the two-stage approach with an image-based proposer to propose and rank hypotheses for difference descriptions. As we do not have ground-truth labels for set differences in this application example, we employ a combination of dataset annotations and external sources, such as public statistics, to validate the correctness of the generated hypotheses (see Fig.\ref{fig:qualitative}). We are able to validate all of the top-8 hypotheses for this example, demonstrating that our proposed approach is able to generate useful difference descriptions in real-world settings with sparse and diluted, a priori unknown differences. 
    
    In this specific example, hypotheses 1, 3, 5, 6, and 8 all relate to significantly higher construction activity in Singapore and the resulting presence of construction workers near roads. In the context of the example application, this could not only prompt us to evaluate how robustly our perception system detects construction workers (which may have been comparatively rare in our training dataset), but also whether the higher density of construction sites introduces additional hazards that must be considered before deploying the fleet in Singapore.

    \section{Summary and Conclusions}
    \label{sec:conclusion}
    This paper studies \emph{object-centric set difference captioning} as a practical dataset introspection tool for autonomous driving.
    By operating on image patches extracted from object detections, the approach avoids confounding scene-level variation and enables aggregation across large collections.
    To support in-domain evaluation, we introduce a benchmark with three dataset splits (web-scraped, annotation-filtered, and CLIP-filtered) and controlled \emph{purity} and \emph{concentration} to model noisy and sparse set differences.
    
    Experiments indicate that autonomous driving data remains substantially more challenging than web imagery, motivating domain-specific evaluation.
    Across methods, a two-stage proposer-to-ranker pipeline consistently provides the most robust behavior under sparsity and noise, as the ranker can aggregate weak signals over the full datasets.
    Among proposers, the image-based and caption-based variants achieve similar accuracy and significantly outperform feature-based proposers.
    
    Accuracy, however, drops sharply for very low concentration and low purity settings, so reliably identifying rare, safety-critical differences remains an open challenge. Future work should therefore focus on improving robustness to extreme sparsity.

\bibliographystyle{IEEEtran}
\bibliography{root}

\end{document}